\documentclass[10pt, a4paper]{article}

\usepackage[a4paper, top=2 cm, bottom=2 cm, left=1.5cm, right=1.5cm]{geometry}
\usepackage[T1]{fontenc}
\usepackage{graphicx}
\usepackage{tikz}
\usetikzlibrary{positioning,arrows.meta,calc,3d,fit,backgrounds,shapes.geometric}

\usepackage[english]{babel}
\usepackage[extrabold]{noto}

\usepackage{amsmath, amsfonts, amssymb} %
\usepackage{booktabs} %
\usepackage{enumitem} %
\usepackage{hyperref} %

\hypersetup{
    colorlinks=false,
    linkcolor=blue,
    filecolor=magenta,
    urlcolor=blue,
    citecolor=blue,
    pdftitle={Vision-Based Density Estimation for Adaptive Traffic Control}, %
}

\makeatletter
\renewcommand\paragraph{\@startsection{paragraph}{4}{\z@}%
                                    {3.25ex \@plus1ex \@minus.2ex}%
                                    {1.5ex \@plus .2ex}%
                                    {\normalfont\normalsize\bfseries}}
\makeatother

\usepackage{authblk}

\usepackage{parskip}              %

\begin{document}
\setlength{\parindent}{0 pt}
\title{\textbf{Vision-Based Density Estimation for Adaptive Traffic Control}}

\author[1]{Mohammad Usaid}
\author[1]{Deepak Silaych}
\author[1,2]{Hrithik Mhatre}
\author[1,*]{Archak Mittal}
\affil[*]{Corresponding Author and Assistant Professor}
\affil[1]{Department of Civil Engineering, Indian Institute of Technology Bombay}
\affil[2]{Department of Computer Science and Engineering, University of California San Diego}
\affil[ ]{\href{mailto:archak@iitb.ac.in}{\texttt{archak@iitb.ac.in}}}
\date{}
\maketitle

\begin{abstract}
\noindent Traffic congestion is becoming a critical challenge in rapidly growing urban centers, particularly in developing nations where rapid urbanization and inadequate transport infrastructure place increasing strain on transport networks. In highly heterogeneous and chaotic traffic flows, camera-based traffic-state estimation is limited by severe vehicle occlusion and overlap. In this paper, we propose a framework to optimize traffic flow directly from stochastic visual observations. The architecture is structurally decoupled into three primary components: (a) vehicle density estimation, (b) predictive dynamics modeling, and (c) policy optimization. The proposed methodology introduces a convolutional neural network (CNN) that learns the mapping from camera feeds to spatial density maps, utilizing an adaptive k-nearest-neighbor (k-NN) Gaussian kernel to handle varying congestion levels. This feeds into an ST3DNet predictive module for spatiotemporal forecasting, followed by a hierarchical Graph-Transformer Multi-Agent Reinforcement Learning (MARL) algorithm to optimize signal phases. The density estimation stage was evaluated on 40 annotated intersection images from Rambagh 4, Jaipur, India, with a reported five-fold validation mean absolute error (MAE) of $3.32 \pm 0.95$ vehicles and a mean absolute percentage error (MAPE) of 14.5\%. These results pertain to the perception component; empirical training and evaluation of the forecasting and control stages remain future work.
\end{abstract}

\section{Introduction}
Rapid global urbanization has placed unprecedented strain on metropolitan transport networks, making the development of ITS a critical priority. This challenge is acutely pronounced in developing countries like India, where urban centers experience extreme traffic congestion. Standard adaptive traffic signal control (ATSC) paradigms, often designed for highly structured, lane-bound, and homogeneous traffic environments typical of Western cities, fail significantly when deployed under chaotic conditions. Indian urban traffic is characterized by extreme heterogeneity. Managing this non-stationary and fluid traffic state requires control systems capable of perceiving spatial density dynamically rather than relying on structured vehicle tracking.

\vspace{0.5em}
Traditional traffic management architectures fail to capture this complexity due to their reliance on localized point-sensor telemetry or idealized probe-vehicle trajectories. Inductive loops are fundamentally constrained by their static placement and their underlying assumption that vehicles occupy discrete, lane-aligned positions. In chaotic traffic flows, vehicles cluster opportunistically, rendering point-sensors highly inaccurate or blind to the true queue dynamics. While high-resolution traffic cameras offer a ubiquitous and non-intrusive alternative filled with rich spatial context, mapping raw, high-dimensional visual streams directly to stable control actions introduces severe optimization challenges, sample inefficiencies, and vulnerability to visual domain shifts.

\section{Related Work}
\subsection{Vision-Based Vehicle Counting and Density Estimation}
Estimating the number and spatial arrangement of vehicles from images is a long-standing problem in traffic monitoring. Guerrero-Gómez-Olmedo \textit{et al.}~\cite{guerrero2015extremely} introduced the TRANCOS benchmark for extremely overlapping vehicles, providing point annotations and a spatial counting evaluation protocol.

Mhatre \textit{et al.}~\cite{mhatre2025pixels} counted moving vehicles by subtracting a background reconstructed through pixel-wise temporal averaging of camera frames, then clustering the foreground into individual vehicles with DBSCAN. Zhang \textit{et al.}~\cite{zhang2016mcnn} proposed a multi-column convolutional neural network (CNN) that maps an image to a density map, using geometry-adaptive Gaussian targets derived from point annotations. This formulation allows a network to be trained on point labels rather than bounding boxes. Although developed for crowd counting, these geometry-adaptive targets provide a methodological precedent for point-supervised density estimation.

Density-based convolutional architectures were subsequently extended to traffic. Li \textit{et al.}~\cite{li2018csrnet} used dilated convolutions to produce high-quality density maps, evaluating on crowd datasets and extending the approach to the TRANCOS vehicle benchmark. Zhang \textit{et al.}~\cite{zhang2017fcn} combined fully convolutional density estimation with recurrent residual count estimation for vehicle counting in low-quality city-camera videos, incorporating temporal information to improve counts. More recently, Jian \textit{et al.}~\cite{jian2025selective} proposed a scale-aware network that produces vehicle density maps and uses optical-flow-based motion for congestion assessment; this work addresses congestion detection rather than signal control.

It is useful to distinguish image-plane count density from physical road density. Hu \textit{et al.}~\cite{hu2021turning} combine vehicle detection with camera calibration to estimate physical density in vehicles per kilometer per lane. Image-plane density maps of the kind produced by the CNN in the proposed framework are count densities: their spatial integral yields a vehicle count, and converting them to physical road density would require geometric calibration.

\subsection{Adaptive Traffic Control and Multi-Agent Coordination}
Deep reinforcement learning has been applied to signal timing, with Li \textit{et al.}~\cite{li2016traffic} demonstrating that a deep neural network could learn a Q-function for signal timing. Subsequent work has refined the learning procedure: Bouktif \textit{et al.}~\cite{bouktifa2023deep} examined consistent state and reward design with double deep Q-learning and prioritized replay in Simulation of Urban MObility (SUMO) simulations. Single-agent formulations can be limited in accounting for cascading effects from adjacent intersections, which motivates coordination across multiple nodes.

To address network effects, researchers have adopted Multi-Agent Reinforcement Learning (MARL). Wu \textit{et al.}~\cite{wu2020multi} proposed a framework using centralized training with decentralized execution, enabling agents to account for neighboring intersections under partial observability. Graph-based architectures have been used to capture spatial dependencies: CoLight~\cite{wei2019colight} employs graph attention for communication among neighboring intersections, and Jiang \textit{et al.}~\cite{jiang2024distributed} introduced a distributed MARL framework using graph decomposition for large-scale signal control. These studies address coordination across intersections, complementing the problem of acquiring traffic-state information from images.

\subsection{Connecting Visual Traffic-State Estimation to Control}
A central question for vision-based control is how visual observations should be converted into the state representation used by a controller. Genders and Razavi~\cite{genders2018evaluating} compared three state resolutions in simulation and found that occupancy and average-speed states performed similarly to individual-vehicle high-resolution states under their setup. This finding cautions against assuming that richer state representations automatically yield better control.

Garg \textit{et al.}~\cite{garg2022vision} proposed an image-based traffic signal controller trained in simulation with domain randomization, and reported simulated control evaluation together with attention visualizations. The real-image experiment examined attention on London traffic-camera images without actuating traffic signals.

The present study investigates spatial vehicle-count density estimation from heterogeneous, occluded intersection imagery, motivated by its potential use as a traffic-state representation for adaptive signal control.

\section{Methodology}
The proposed framework (Figure~\ref{fig:framework}) presents a vision-based framework for ATSC. The controller state combines current and forecast density representations with signal-phase information and relevant masks. To support forecasting and signal-control decisions, the system architecture is structurally decoupled into three computational modules:
\begin{enumerate}
\item State Estimation, utilizing a convolutional neural network to regress raw RGB camera feeds into continuous spatial density distributions, with the architecture following Tian \textit{et al.}~\cite{tian2019automated}; overlapping-vehicle counting provides the motivating context~\cite{guerrero2015extremely}.
\item Predictive Dynamics Modeling, employing a proposed ST3DNet-based module to forecast future spatial density maps from historical density sequences~\cite{guo2019deep}.
\item Policy Optimization, wherein a proposed hierarchical graph-based MARL controller uses these density representations to select joint signal phase configurations across the traffic network.
\end{enumerate}
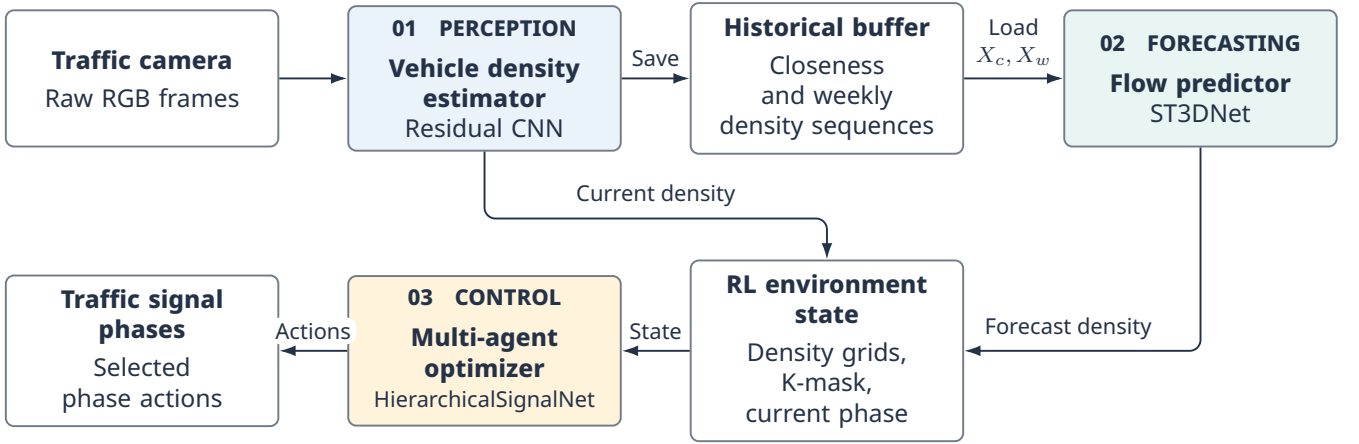
\begin{figure}[!htbp]
\centering
\resizebox{\textwidth}{!}{%
\definecolor{ovInk}{HTML}{243247}
\definecolor{ovBlue}{HTML}{EAF2FA}
\definecolor{ovTeal}{HTML}{E9F5F2}
\definecolor{ovAmber}{HTML}{FFF3DB}
\begin{tikzpicture}[
 font=\sffamily\fontsize{9}{11}\selectfont,text=ovInk,
 ovbox/.style={draw=ovInk!65,line width=.7pt,rounded corners=3pt,align=center,text width=3.15cm,minimum height=1.75cm,inner sep=5pt,fill=white},
 ovarr/.style={-{Latex[length=2.1mm,width=1.4mm]},draw=ovInk,line width=.7pt,rounded corners=4pt},
 ovlab/.style={font=\sffamily\fontsize{8}{10}\selectfont,text=ovInk,fill=white,inner sep=2pt}]
\node[ovbox] (camera) at (0,0) {\textbf{Traffic camera}\\[3pt]Raw RGB frames};
\node[ovbox,fill=ovBlue] (s1) at (4.4,0) {{\fontsize{8}{10}\selectfont\bfseries 01\quad PERCEPTION}\\[4pt]\textbf{Vehicle density\\estimator}\\Residual CNN};
\node[ovbox] (history) at (8.8,0) {\textbf{Historical buffer}\\[3pt]Closeness and weekly\\density sequences};
\node[ovbox,fill=ovTeal] (s2) at (13.6,0) {{\fontsize{8}{10}\selectfont\bfseries 02\quad FORECASTING}\\[4pt]\textbf{Flow predictor}\\ST3DNet};
\node[ovbox] (state) at (8.8,-3.5) {\textbf{RL environment\\state}\\[3pt]Density grids, K-mask,\\current phase};
\node[ovbox,fill=ovAmber] (s3) at (4.4,-3.5) {{\fontsize{8}{10}\selectfont\bfseries 03\quad CONTROL}\\[4pt]\textbf{Multi-agent optimizer}\\{\fontsize{8}{10}\selectfont HierarchicalSignalNet}};
\node[ovbox] (out) at (0,-3.5) {\textbf{Traffic signal\\phases}\\[3pt]Selected phase actions};
\draw[ovarr] (camera)--(s1);
\draw[ovarr] (s1)--node[ovlab,above=2pt]{Save}(history);
\draw[ovarr] (history)--node[ovlab,above=2pt,align=center]{Load\\$X_c,X_w$}(s2);
\draw[ovarr] (s1.south)--(4.4,-1.8)--node[ovlab,above=2pt]{Current density}(8.8,-1.8)--(state.north);
\draw[ovarr] (s2.south)|-node[ovlab,pos=.78,above=2pt]{Forecast density}(state.east);
\draw[ovarr] (state)--node[ovlab,above=2pt]{State}(s3);
\draw[ovarr] (s3)--node[ovlab,above=2pt]{Actions}(out);
\end{tikzpicture}%
}
\caption{Three-stage pipeline. Current density and forecast density enter the controller state; the controller selects traffic-signal phases.}
\label{fig:framework}
\end{figure}

\subsection{Dataset \& Preprocessing}
\subsubsection{Data Acquisition and Point Annotation}
The raw visual data was collected from the RAMBAGH 4 intersection in Jaipur, India. The dataset comprises 40 high-resolution ($720\times480$ pixels) RGB frames captured by a standard traffic surveillance camera. To ensure the dataset captures statistically independent traffic states and a high variance of kinematic conditions, the frames were sampled at 30-second intervals.

\begin{figure}[!htbp]
\centering
\includegraphics[width=\textwidth]{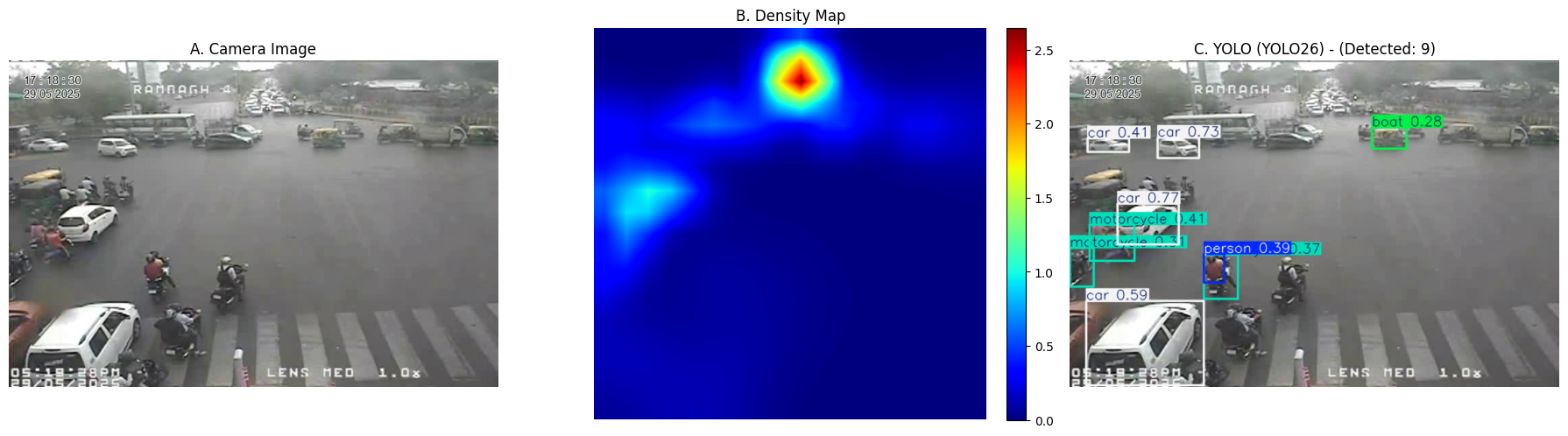}
\caption{Example camera frame (A), annotation-derived density map (B), and YOLO26 detections (C). This illustration is not a benchmark comparison.}
\label{fig:yolo-failure}
\end{figure}

Indian urban traffic is characterized by extreme heterogeneity, the absence of strict lane discipline, and the presence of non-standard vehicle classes, leading to vehicles occupying opportunistic, highly dense clusters. To validate the limitations of standard object detection paradigms in this domain, we applied a state-of-the-art bounding-box detector (YOLO)~\cite{jocher2026yolo26} to our dataset (Figure~\ref{fig:yolo-failure}). The discrete detector struggled severely with the occlusion and low-resolution telemetry, failing to capture the true volume of the traffic (detecting merely 9 objects) while exhibiting catastrophic misclassifications, most notably identifying a distant vehicle as a ``boat.'' These failures definitively underscore why discrete bounding-box tracking is brittle and computationally prohibitive for Indian traffic monitoring.

\vspace{0.5em}
Therefore, a dotting annotation strategy was employed using the VGG Image Annotator~\cite{dutta2019via}. For each frame, a single 2D coordinate point $(cx, cy)$ was manually annotated at the geometric center of every visible vehicle. This approach cleanly bypasses the limitations of bounding-box overlap and algorithmic misclassification, providing the exact ground-truth count and spatial topology required to generate density maps.

\subsubsection{Adaptive k-NN Spatial Density Generation}
Generating a vehicle-density map involves smoothing each annotated point with a 2D Gaussian kernel. A fixed bandwidth applies the same spread throughout the image, whereas geometry-adaptive kernels vary the spread with neighbouring-point spacing~\cite{zhang2016mcnn}. Here, an adaptive k-nearest-neighbour (k-NN) Gaussian kernel is used.

\vspace{0.5em}
First, each annotated point $(cx, cy)$ is projected from the original image dimensions $(W, H)$ to the target density map dimensions $(W', H')$:
$$mx = cx \left( \frac{W'}{W} \right), \quad my = cy \left( \frac{H'}{H} \right)$$

\vspace{0.5em}
The bandwidth of the Gaussian kernel for each vehicle $i$ is determined by its proximity to surrounding vehicles. A k-dimensional tree (KD-tree) is used to query the $k = 3$ nearest neighbouring vehicles. Their average distance, denoted as $\bar{d}_{k\text{-NN}}^i$, characterizes local point spacing. The bandwidth $\sigma_i$ is set to half this average distance and clipped between 0.8 and 2.5 to limit excessively narrow or broad kernels:
$$\sigma_i = \max \left(0.8, \min(0.5 \times \bar{d}_{k\text{-NN}}^i, 2.5) \right)$$

\vspace{0.5em}
For each projected coordinate $(mx, my)$, a unit impulse is placed on a zero-initialized discrete grid at the nearest integer index. A Gaussian filter with the corresponding bandwidth $\sigma_i$ is then applied. To preserve each annotated vehicle's contribution despite any boundary truncation, its density map $d_i(p)$ is normalized so that its spatial sum equals one within numerical precision:
$$\hat{d}_i(p) = \frac{d_i(p)}{\sum_{p} d_i(p)}$$
The complete density map $\mathcal{D}^*$ is constructed by summing these normalized maps for all $N$ annotated vehicles:
$$\mathcal{D}^*(p) = \sum_{i=1}^{N} \hat{d}_i(p)$$

\vspace{0.5em}
A final global scaling operation corrects residual numerical discrepancies by multiplying the density map by the ratio of the annotated vehicle count to its current sum:
$$\mathcal{D}^*_{final} = \mathcal{D}^* \left( \frac{N}{\sum_{p} \mathcal{D}^*(p)} \right)$$
The resulting $18\times18$ map sums to the annotated count within numerical precision and approximately represents the annotated spatial locations. This description applies to the dataset frames, which contain 12--49 annotated vehicles.

\subsection{Problem Formulation}
The proposed multi-intersection traffic signal control task is formulated as a cooperative traffic-control problem under partial observability, modeled at the network level as a Partially Observable Markov Decision Process (POMDP). Let $\mathcal{V} = \{1, 2, \dots, N\}$ represent the set of controllable intersections in the road network. Decisions are taken at discrete index $t \in \{0, 1, \dots, T-1\}$ over a finite horizon of $T$ decisions, with a fixed control interval $\Delta t$; the physical time corresponding to decision index $t$ is $t\Delta t$.

\subsubsection{Decision Variables}
At each decision index $t$, the controller selects or retains a green-phase index for each intersection $i \in \mathcal{V}$, subject to the timing constraints below. Let $\Phi^i$ denote the set of all valid green phases for intersection $i$. The action is the phase index
$$a_t^i \in \mathcal{A}^i = \{0, 1, \dots, |\Phi^i| - 1\} \quad \forall i \in \mathcal{V}$$
The joint selected action tuple is $\mathbf{A}_t = (a_t^1, a_t^2, \dots, a_t^N)$, and the joint action space is the Cartesian product $\mathcal{A} = \mathcal{A}^1 \times \mathcal{A}^2 \times \dots \times \mathcal{A}^N$.

\subsubsection{Objective Function}
The objective is to reduce vehicle waiting at the controlled intersections. For the proposed reward, let $W_t^i$ denote the sum of accumulated waiting times of vehicles currently in the monitored approaches of intersection $i$ at decision $t$. The base reward is defined as the negative difference in this quantity between consecutive decisions:
$$R_t^{\text{base}} = - \sum_{i=1}^{N} \left( W_{t+1}^i - W_t^i \right)$$
In the proposed Stage~3 design, this base reward is extended by fairness and coordination terms; the full reward is denoted $R_t$. The optimization objective is to find a joint policy $\pi_\theta$ (parameterized by neural network weights $\theta$) that maximizes the expected cumulative discounted full reward:
$$\max_\theta \mathcal{J}(\pi_\theta) = \mathbb{E}_{\tau \sim \pi_\theta} \left[ \sum_{t=0}^{T-1} \gamma^t R_t \right]$$
where $\gamma \in [0, 1)$ is the discount factor, $\mathbf{S}_t$ is the hidden environment state, and $\tau$ represents the trajectory of the environment.

\subsubsection{System Constraints}
The proposed formulation includes operational constraints on green-phase timing.
\paragraph{Minimum Green Time Constraint}
To limit rapid phase oscillation, an active green phase is held for a minimum green duration $g_{min}$. When a green phase is active, let $\tau_t^i$ be the elapsed green duration since the current green phase began, excluding yellow clearance, and let $q_t^i$ be the currently active green phase index. The admissible action set is
$$\mathcal{A}_t^i = \begin{cases}
    \{q_t^i\}, & \text{if } \tau_t^i < g_{min} \\
    \mathcal{A}^i, & \text{otherwise}
    \end{cases} \qquad a_t^i \in \mathcal{A}_t^i$$
During yellow clearance, the previously selected target green is retained and further phase changes are masked. When the selected green phase differs from $q_t^i$, the environment inserts a yellow clearance interval $y$ and then activates the target green. Minimum-green masking is proposed before action selection, while the physical transition is to be enforced by the environment.

\paragraph{State Transition Dynamics}
The transition to the next hidden state $\mathbf{S}_{t+1}$ follows the underlying kinematic models of the traffic simulator (car-following and lane-changing models):
$$\mathbf{S}_{t+1} \sim \mathcal{P}(\cdot | \mathbf{S}_t, \mathbf{A}_t)$$

\paragraph{State Observability Constraint}
The hidden state $\mathbf{S}_t$ is not directly observed. Instead, the policy uses local observations $o_t^i$ comprising current and historical estimated density maps, forecast density maps, phase information and relevant masks. Stage~1 supplies the estimated density maps, Stage~2 supplies the forecasts, and the signal controller supplies the phase and timing information used in the masks. The proposed graph-based controller also uses messages derived from observations at connected intersections.

\subsection{Stage 1: Vehicle Density Estimation Model}
\label{sec:stage1}
\begin{figure}[!htbp]
\centering
\resizebox{\linewidth}{!}{%
\begingroup\sffamily
\definecolor{dOneInk}{HTML}{243247}
\definecolor{dOneBlue}{HTML}{EAF2FA}
\definecolor{dOneTeal}{HTML}{E9F5F2}
\definecolor{dOneAmber}{HTML}{FFF3DB}
\definecolor{dOneViolet}{HTML}{F0ECFA}
\begin{tikzpicture}[
 font=\sffamily\fontsize{10}{12}\selectfont,text=dOneInk,
 d1box/.style={draw=dOneInk!65,line width=.65pt,rounded corners=2.5pt,minimum height=1.03cm,text width=2.8cm,align=center,inner sep=4pt,fill=white},
 d1conv/.style={d1box,fill=dOneBlue},
 d1pool/.style={d1box,fill=black!3},
 d1dw/.style={d1box,fill=dOneTeal},
 d1head/.style={d1box,fill=dOneAmber},
 d1io/.style={d1box,fill=white,draw=dOneInk,minimum height=.78cm},
 d1arr/.style={-{Latex[length=2.1mm,width=1.4mm]},draw=dOneInk,line width=.7pt,rounded corners=3pt},
 d1sum/.style={circle,draw=dOneInk,line width=.8pt,fill=white,minimum size=.53cm,inner sep=0pt,font=\sffamily\bfseries\large},
 d1title/.style={font=\sffamily\bfseries\fontsize{9}{11}\selectfont,text=dOneInk,align=center},
 d1small/.style={font=\sffamily\fontsize{10}{12}\selectfont,text=dOneInk,align=center}]
\node[d1title] (h1) at (0,1.7) {01\quad Convolutional stem};
\node[d1title] (h2) at (7.1,1.7) {02\quad Residual block 1};
\node[d1title] (h3) at (13.45,1.7) {03\quad Residual block 2};
\node[d1title] (h4) at (18.7,1.7) {04\quad Regression head};
\node[d1io] (inp) at (0,.45) {\textbf{RGB input\\image}\\[-1pt]{\fontsize{10}{12}\selectfont $[3,H,W]$}};
\node[d1conv] (s1) at (0,-1.05) {\textbf{Conv $7\times7$}\\[-1pt]{\fontsize{10}{12}\selectfont $[32,H,W]$}};
\node[d1pool] (s2) at (0,-2.65) {\textbf{MaxPool $2\times2$}\\[-1pt]{\fontsize{10}{12}\selectfont $[32,H/2,W/2]$}};
\node[d1conv] (s3) at (0,-4.25) {\textbf{Conv $7\times7$}\\[-1pt]{\fontsize{10}{12}\selectfont $[32,H/2,W/2]$}};
\node[d1pool] (s4) at (0,-5.85) {\textbf{MaxPool $3\times3$}\\[-1pt]{\fontsize{10}{12}\selectfont $[32,H/2,W/2]$}};
\foreach \a/\b in {inp/s1,s1/s2,s2/s3,s3/s4}{\draw[d1arr] (\a)--(\b);}
\coordinate (r1in) at (5.8,.45);
\node[d1conv] (r11) at (5.8,-1.05) {\textbf{Conv $1\times1$}\\[-1pt]{\fontsize{10}{12}\selectfont $[32,H/2,W/2]$}};
\node[d1dw] (r12) at (5.8,-2.65) {\textbf{Conv $3\times3$}\quad Groups=32\\[-1pt]{\fontsize{10}{12}\selectfont $[32,H/2,W/2]$}};
\node[d1conv] (r13) at (5.8,-4.25) {\textbf{Conv $1\times1$}\\[-1pt]{\fontsize{10}{12}\selectfont $[32,H/4,W/4]$}};
\node[d1box,fill=dOneViolet,text width=2.4cm,minimum height=1.35cm] (proj) at (8.8,-2.65) {\textbf{Conv $2\times2$}\\Projection shortcut\\[-1pt]{\fontsize{10}{12}\selectfont $[32,H/4,W/4]$}};
\node[d1sum] (sum1) at (5.8,-5.85) {$+$};
\draw[d1arr] (r1in)--(r11);
\foreach \a/\b in {r11/r12,r12/r13,r13/sum1}{\draw[d1arr] (\a)--(\b);}
\draw[d1arr] (r1in)-|(proj.north);
\draw[d1arr] (proj.south)|-(sum1.east);
\fill[dOneInk] (r1in) circle[radius=1.1pt];
\coordinate (r2in) at (12.3,.45);
\node[d1conv] (r21) at (12.3,-1.05) {\textbf{Conv $1\times1$}\\[-1pt]{\fontsize{10}{12}\selectfont $[32,H/4,W/4]$}};
\node[d1dw] (r22) at (12.3,-2.65) {\textbf{Conv $3\times3$}\quad Groups=32\\[-1pt]{\fontsize{10}{12}\selectfont $[32,H/4,W/4]$}};
\node[d1conv] (r23) at (12.3,-4.25) {\textbf{Conv $1\times1$}\\[-1pt]{\fontsize{10}{12}\selectfont $[32,H/4,W/4]$}};
\node[d1sum] (sum2) at (12.3,-5.85) {$+$};
\draw[d1arr] (r2in)--(r21);
\foreach \a/\b in {r21/r22,r22/r23,r23/sum2}{\draw[d1arr] (\a)--(\b);}
\draw[d1arr] (r2in)--(14.85,.45)--(14.85,-5.85)--(sum2.east);
\node[d1small,fill=black!1,inner sep=4pt,text width=1.4cm] at (14.85,-2.65) {Identity\\shortcut};
\fill[dOneInk] (r2in) circle[radius=1.1pt];
\node[d1head] (reg1) at (18.7,-1.05) {\textbf{Conv $5\times5$}\\[-1pt]{\fontsize{10}{12}\selectfont $[64,H/4,W/4]$}};
\node[d1head] (reg2) at (18.7,-2.65) {\textbf{Conv $1\times1$}\\[-1pt]{\fontsize{10}{12}\selectfont $[1000,H/4,W/4]$}};
\node[d1head] (reg3) at (18.7,-4.25) {\textbf{Conv $1\times1$}\\[-1pt]{\fontsize{10}{12}\selectfont $[400,H/4,W/4]$}};
\node[d1head] (reg4) at (18.7,-5.85) {\textbf{Conv $1\times1$}\\[-1pt]{\fontsize{10}{12}\selectfont $[1,H/4,W/4]$}};
\node[d1io] (out) at (18.7,-7.5) {\textbf{Predicted density map}\\[-1pt]{\fontsize{10}{12}\selectfont $[1,H/4,W/4]$}};
\foreach \a/\b in {reg1/reg2,reg2/reg3,reg3/reg4,reg4/out}{\draw[d1arr] (\a)--(\b);}
\draw[d1arr] (s4.south)--(0,-6.95)--(3.1,-6.95)--(3.1,.45)--(r1in);
\draw[d1arr] (sum1.south)--(5.8,-6.95)--(10.5,-6.95)--(10.5,.45)--(r2in);
\draw[d1arr] (sum2.south)--(12.3,-6.95)--(16.3,-6.95)--(16.3,.45)--(18.7,.45)--(reg1.north);
\begin{scope}[on background layer]
\foreach \xmin/\xmax in {-2.0/2.0,4.0/10.15,10.9/15.75,16.75/20.65}{
\draw[draw=dOneInk!18,fill=black!1,rounded corners=5pt,line width=.6pt] (\xmin,2.15) rectangle (\xmax,-6.6);}
\end{scope}
\node[d1small,anchor=west] at (-2,-8.65) {\textbf{Notation}\quad $[C,H,W]$: channels and spatial dimensions.};
\node[d1small,anchor=west] at (-2,-9.15) {Blue: standard Conv2D + batch normalization + ReLU\quad Teal: depthwise Conv2D};
\end{tikzpicture}
\endgroup%
}
\caption{Vehicle density estimator: convolutional stem, two residual blocks, and a $32\rightarrow64\rightarrow1000\rightarrow400\rightarrow1$ regression head.}
\label{fig:stage-1}
\end{figure}
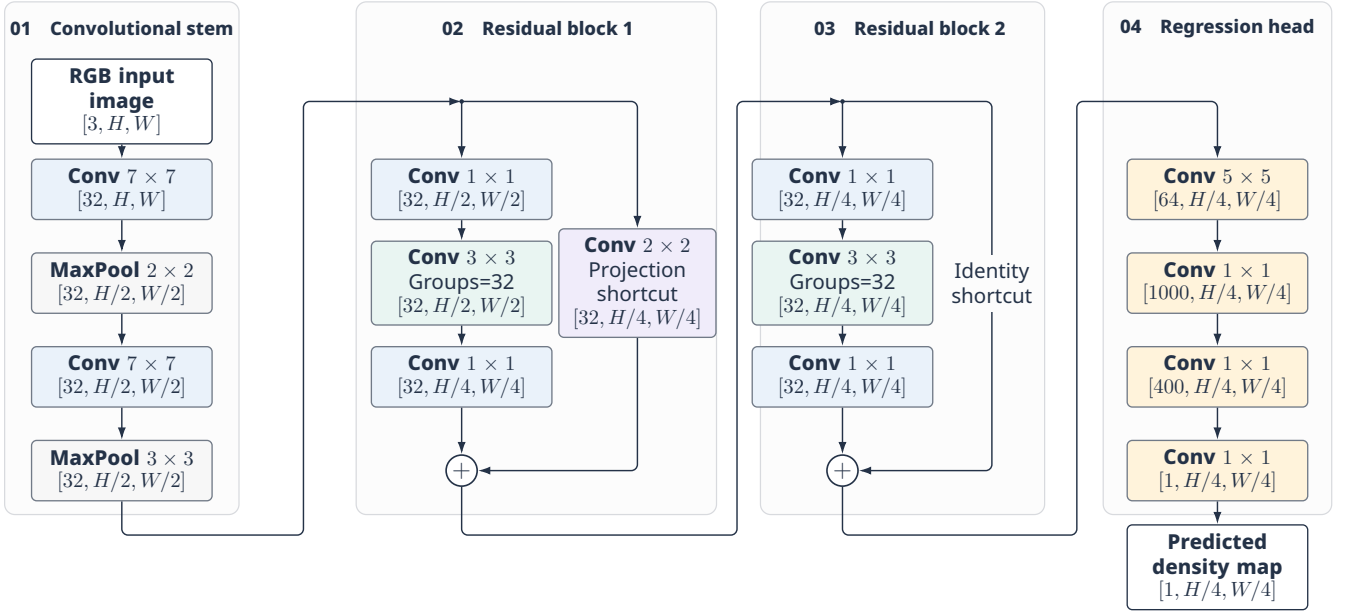

To perceive the current traffic state from unstructured visual data, the vehicle density estimation task (Figure~\ref{fig:stage-1}) is formulated as a supervised Empirical Risk Minimization (ERM) problem. Each camera frame is an RGB image of original resolution $720\times480$, resized to $72\times72$ with pixel values divided by 255 to the range $[0,1]$. Let the input be $I \in \mathbb{R}^{3 \times H \times W}$ with $H = W = 72$. The objective is to learn a non-linear mapping function $F_\omega: I \rightarrow \hat{\mathcal{D}}$, where $\hat{\mathcal{D}} \in \mathbb{R}^{1 \times H' \times W'}$ is the predicted discrete spatial density map with $H' = W' = 18$, and $\omega \in \Omega$ represents the learnable parameters of the convolutional neural network (CNN) within the parameter space $\Omega$.

\subsubsection{Optimization Objective}
The optimal parameter set $\omega^*$ is obtained by minimizing a pixel-mean squared density error plus a weighted squared count error over $M$ training samples. Let $\hat{\mathcal{D}}^{(m)} = F_\omega(I^{(m)})$ be the predicted density map for sample $m$, $\mathcal{D}^{*(m)}$ its target density map, and $N_m = \sum_{u,v} \mathcal{D}^{*(m)}_{u,v}$ its annotated vehicle count:

$$\omega^* = \arg\min_\omega \frac{1}{M} \sum_{m=1}^{M} \Bigg[ \frac{1}{H'W'} \sum_{u=1}^{H'} \sum_{v=1}^{W'} \left( \hat{\mathcal{D}}^{(m)}_{u,v} - \mathcal{D}^{*(m)}_{u,v} \right)^2  + \lambda_c \left( \sum_{u=1}^{H'} \sum_{v=1}^{W'} \hat{\mathcal{D}}^{(m)}_{u,v} - N_m \right)^2 \Bigg]$$
with count weight $\lambda_c = 2$.

\subsubsection{Architectural Mapping and Structural Constraints}
The architecture in Figure~\ref{fig:stage-1} follows Tian \textit{et al.}~\cite{tian2019automated} and consists of a convolutional stem, two residual blocks, and a regression head. The stem alternates two $7\times7$ convolutions with max-pooling operations. The first $2\times2$ pooling layer halves the spatial dimensions, while the second $3\times3$ pooling layer uses stride 1 and preserves them; the stem outputs 32 channels. The first residual block downsamples by a further factor of two through its final $1 \times 1$ convolution and a $2 \times 2$ projection shortcut. The second residual block retains the reduced spatial resolution and uses an identity shortcut.

\vspace{0.5em}
The residual blocks use $3 \times 3$ depthwise convolutions with 32 groups and 32 input channels. For a given output channel $c$, the depthwise convolution is computed over the singleton input-channel group $\mathcal{G}_c = \{c\}$:
$$\mathbf{Y}_c = \sum_{i \in \mathcal{G}_c} \mathbf{K}_{c,i} * \mathbf{X}_i + b_c$$
Here, $\mathbf{X}_i$ is the $i$-th input feature map, $\mathbf{K}_{c,i}$ represents the spatial convolution kernel mapping input $i$ to output $c$, $b_c$ is the channel-specific bias, and $*$ denotes the 2D cross-correlation operator. Dropout2d with $p=0.25$ is applied after each residual block.

\vspace{0.5em}
The regression head produces the density map through successive convolutional projections with channel dimensions $32 \rightarrow 64 \rightarrow 1000 \rightarrow 400 \rightarrow 1$ and kernel sizes $5 \times 5, 1 \times 1, 1 \times 1, 1 \times 1$. Dropout2d with $p=0.25$ is applied after each of the first two head blocks, and a final rectified linear unit (ReLU) yields nonnegative density values.

\vspace{0.5em}
The total estimated vehicle count $\hat{N}$ within a given frame is computed as the spatial sum over the generated density map:
$$\hat{N} = \sum_{u=1}^{H'} \sum_{v=1}^{W'} \hat{\mathcal{D}}_{u,v}$$
This sum estimates the annotated vehicle count. The spatial density map $\hat{\mathcal{D}}$ is retained as an intended input to the proposed Stage~2 predictive dynamics module.

\subsection{Stage 2: Density Map Forecasting}
\label{sec:stage2}
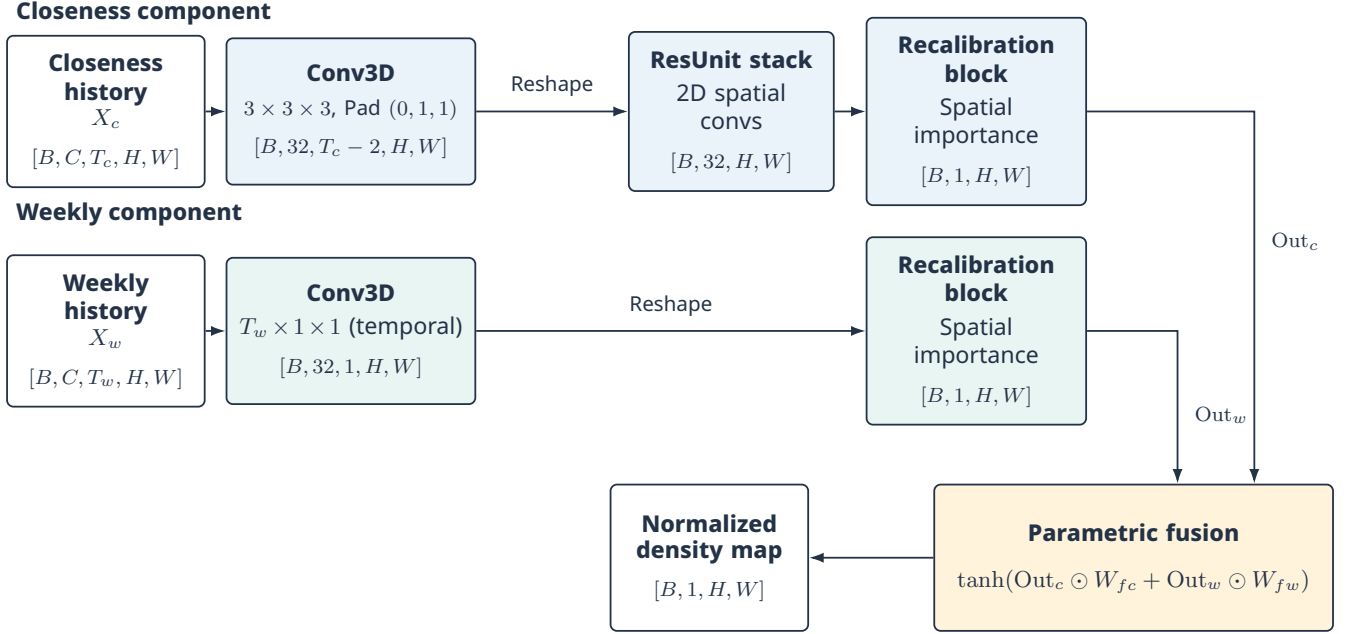
\begin{figure}[!htbp]
\centering
\resizebox{\linewidth}{!}{%
\begingroup\sffamily
\begingroup
\definecolor{sttwoNavy}{HTML}{243247}
\definecolor{sttwoBlue}{HTML}{EAF2FA}
\definecolor{sttwoTeal}{HTML}{E9F5F2}
\definecolor{sttwoAmber}{HTML}{FFF3DB}
\begin{tikzpicture}[
  x=1cm,y=1cm,
  font=\sffamily\fontsize{9}{11}\selectfont,
  text=sttwoNavy,
  >={Latex[length=2mm,width=1.4mm]},
  sttwo box/.style={
    draw=sttwoNavy,line width=0.7pt,rounded corners=2.5pt,
    align=center,inner xsep=5pt,inner ysep=7pt,
    minimum height=2.0cm,fill=white
  },
  sttwo input/.style={sttwo box,text width=2.35cm,minimum height=2.0cm},
  sttwo conv/.style={sttwo box,text width=3.05cm},
  sttwo residual/.style={sttwo box,text width=2.45cm},
  sttwo recal/.style={sttwo box,text width=2.65cm},
  sttwo fusion/.style={sttwo box,text width=5.10cm,minimum height=2.0cm,fill=sttwoAmber},
  sttwo output/.style={sttwo box,text width=2.35cm,minimum height=2.0cm},
  sttwo arrow/.style={draw=sttwoNavy,line width=0.7pt,-{Latex[length=2mm,width=1.4mm]}},
  sttwo tensor/.style={font=\sffamily\fontsize{8}{10}\selectfont},
  sttwo branch/.style={anchor=west,font=\sffamily\bfseries\fontsize{9}{11}\selectfont},
  sttwo edge label/.style={fill=white,inner sep=2.5pt,font=\sffamily\fontsize{8}{10}\selectfont}
]

\node[sttwo branch] at (0.00,3.65) {Closeness component};
\node[sttwo branch] at (0.00,0.90) {Weekly component};

\node[sttwo input] (sttwo-in-c) at (1.35,2.30) {
  \textbf{Closeness history}\\[-1pt]
  $X_c$\\[4pt]
  {\fontsize{8}{10}\selectfont $[B,C,T_c,H,W]$}
};
\node[sttwo conv,fill=sttwoBlue] (sttwo-conv-c) at (4.70,2.30) {
  \textbf{Conv3D}\\[2pt]
  {\fontsize{8}{10}\selectfont $3\times3\times3$, Pad $(0,1,1)$}\\[4pt]
  {\fontsize{8}{10}\selectfont $[B,32,T_c-2,H,W]$}
};
\node[sttwo residual,fill=sttwoBlue] (sttwo-res-c) at (9.90,2.30) {
  \textbf{ResUnit stack}\\[2pt]
  2D spatial\\convs\\[4pt]
  {\fontsize{8}{10}\selectfont $[B,32,H,W]$}
};
\node[sttwo recal,fill=sttwoBlue] (sttwo-recal-c) at (13.25,2.30) {
  \textbf{Recalibration block}\\[2pt]
  Spatial\\importance\\[4pt]
  {\fontsize{8}{10}\selectfont $[B,1,H,W]$}
};

\node[sttwo input] (sttwo-in-w) at (1.35,-0.70) {
  \textbf{Weekly history}\\[-1pt]
  $X_w$\\[4pt]
  {\fontsize{8}{10}\selectfont $[B,C,T_w,H,W]$}
};
\node[sttwo conv,fill=sttwoTeal] (sttwo-conv-w) at (4.70,-0.70) {
  \textbf{Conv3D}\\[2pt]
  $T_w\times1\times1$ (temporal)\\[4pt]
  {\fontsize{8}{10}\selectfont $[B,32,1,H,W]$}
};
\node[sttwo recal,fill=sttwoTeal] (sttwo-recal-w) at (13.25,-0.70) {
  \textbf{Recalibration block}\\[2pt]
  Spatial\\importance\\[4pt]
  {\fontsize{8}{10}\selectfont $[B,1,H,W]$}
};

\node[sttwo fusion] (sttwo-fusion) at (15.40,-3.80) {
  \textbf{Parametric fusion}\\[7pt]
  $\tanh\!\left(\mathrm{Out}_c\odot W_{fc}
  +\mathrm{Out}_w\odot W_{fw}\right)$
};
\node[sttwo output] (sttwo-out) at (9.60,-3.80) {
  \textbf{Normalized}\\[-1pt]
  \textbf{density map}\\[4pt]
  {\fontsize{8}{10}\selectfont $[B,1,H,W]$}
};

\draw[sttwo arrow] (sttwo-in-c.east) -- (sttwo-conv-c.west);
\draw[sttwo arrow] (sttwo-conv-c.east) --
  node[sttwo edge label,above=3pt] {Reshape} (sttwo-res-c.west);
\draw[sttwo arrow] (sttwo-res-c.east) -- (sttwo-recal-c.west);
\draw[sttwo arrow] (sttwo-in-w.east) -- (sttwo-conv-w.west);
\draw[sttwo arrow] (sttwo-conv-w.east) --
  node[sttwo edge label,above=3pt] {Reshape} (sttwo-recal-w.west);

\coordinate (sttwo-c-in) at ([xshift=1.65cm]sttwo-fusion.north);
\coordinate (sttwo-w-in) at ([xshift=0.60cm]sttwo-fusion.north);
\draw[sttwo arrow] (sttwo-recal-c.east) --
  (sttwo-c-in |- sttwo-recal-c.east) --
  node[sttwo tensor,right=3pt,pos=0.35] {$\mathrm{Out}_c$} (sttwo-c-in);
\draw[sttwo arrow] (sttwo-recal-w.east) --
  (sttwo-w-in |- sttwo-recal-w.east) --
  node[sttwo tensor,right=3pt,pos=0.55] {$\mathrm{Out}_w$} (sttwo-w-in);
\draw[sttwo arrow] (sttwo-fusion.west) -- (sttwo-out.east);

\end{tikzpicture}
\endgroup
\endgroup%
}
\caption{ST3DNet-based forecasting architecture with closeness and weekly branches, spatial recalibration, and parametric fusion.}
\label{fig:stage-2}
\end{figure}

Traffic patterns exhibit strong periodic and trend-based temporal dependencies. We formulate the density-map forecasting task (Figure~\ref{fig:stage-2}) as an optimization problem designed to map historical density maps into future spatial distributions. The inputs are single-channel density maps produced by Stage~1, with $C = 1$ and $H = W = 18$; in Figure~\ref{fig:stage-2} these are processed in batches of size $B$. Let $\Delta t$ denote the sampling interval between consecutive maps and $h$ a positive integer horizon in steps. For normalization, we propose a fixed positive density scale $s$, with $s > 0$ shared across all input and target maps and fixed at inference, chosen from training data only: $Z = 2\mathcal{D}/s - 1$, where $0 \leq \mathcal{D}_{u,v} \leq s$, $\mathcal{D}$ denotes physical count-density, and $Z$ denotes normalized values. The input consists of two historical sequences of normalized density maps, both available at sampling index $t$: a ``closeness'' sequence $\mathcal{X}_c \in \mathbb{R}^{T_c \times H \times W}$ stacking the normalized maps of the last $T_c$ consecutive sampling indices through $t$, and a ``weekly'' sequence $\mathcal{X}_w \in \mathbb{R}^{T_w \times H \times W}$ stacking normalized maps aligned with the target time from previous $T_w$ weeks and available by $t\Delta t$, with $T_c \geq 3$ and $T_w \geq 1$. The target index is $t + h$, corresponding to physical time $(t + h)\Delta t$. The objective is to learn an optimal parameterization $\psi^*$ for the predictive mapping $\Psi_\psi : (\mathcal{X}_c, \mathcal{X}_w) \rightarrow \hat{Z}_{future}$ that minimizes the empirical forecasting risk over $M$ training examples:

$$\psi^* = \arg\min_\psi \frac{1}{M} \sum_{m=1}^{M} \mathcal{L}_{forecast} \left( \Psi_\psi(\mathcal{X}_c^{(m)}, \mathcal{X}_w^{(m)}), Z_{future}^{*(m)} \right)$$

where $Z_{future}^{*(m)}$ is the normalized target and the forecast loss is the mean squared pixel error over $H \times W$:
$$\mathcal{L}_{forecast}(A, Q) = \frac{1}{HW} \sum_{u=1}^{H} \sum_{v=1}^{W} \left( A_{u,v} - Q_{u,v} \right)^2$$

The architectural mapping $\Psi_\psi$ uses a 3D Spatio-Temporal Network (ST3DNet)~\cite{guo2019deep} that processes the temporal streams independently before executing a parametric fusion.

\vspace{0.5em}
To capture immediate kinematic dependencies, $\mathcal{X}_c$ is subjected to a 3D convolution with 32 filters and a $(3 \times 3 \times 3)$ kernel with spatial padding $(0, 1, 1)$. Zero-padding is omitted from the temporal dimension, shrinking the temporal depth from $T_c$ to $T_c - 2$, giving a tensor of shape $B \times 32 \times (T_c - 2) \times H \times W$. The tensor is subsequently reshaped by merging the temporal and channel axes into a 2D spatial representation $\mathcal{G}_c \in \mathbb{R}^{K_c \times H \times W}$, where $K_c = 32(T_c - 2)$. This representation is refined via a stack of $U \geq 1$ Residual Units. The first unit maps $K_c$ to 32 channels with a matching projection shortcut when needed, and the remaining units preserve 32 channels, yielding $\mathcal{F}_c \in \mathbb{R}^{32 \times H \times W}$. The batch dimension is omitted in these per-example tensors.

\vspace{0.5em}
To model macro-level recurrent patterns, $\mathcal{X}_w$ undergoes a temporal-collapsing 3D convolution with 32 filters and a $(T_w \times 1 \times 1)$ kernel, producing a tensor of shape $B \times 32 \times 1 \times H \times W$ that is reshaped into $\mathcal{F}_w \in \mathbb{R}^{32 \times H \times W}$.

\vspace{0.5em}
Because intersection topologies are heterogeneous, we use a Recalibration Block to reweight spatial importance. Both intermediate tensors $\mathcal{F}_c$ and $\mathcal{F}_w$ are independently recalibrated using separate learned weight matrices $\mathbf{W}_{rc,c}$ and $\mathbf{W}_{rc,w} \in \mathbb{R}^{32 \times H \times W}$, fixed at inference, by computing the Hadamard product ($\odot$) and integrating across the feature channels $k$:

$$Y_b = \sum_{k=1}^{32} \left( \mathcal{F}_{b,k} \odot \mathbf{W}_{rc,b,k} \right), \quad b \in \{c, w\}$$

This squashes the multi-channel feature maps into single-channel representations $Y_c, Y_w \in \mathbb{R}^{1 \times H \times W}$.

\vspace{0.5em}
Finally, the future normalized density map is synthesized via a parametric fusion. We define fusion weight matrices $\mathbf{W}_{fc}$ and $\mathbf{W}_{fw} \in \mathbb{R}^{1 \times H \times W}$, learned and fixed at inference, that scale the closeness and weekly recalibrated representations. The output is bounded using a hyperbolic tangent activation:

$$\hat{Z}_{future} = \tanh(Y_c \odot \mathbf{W}_{fc} + Y_w \odot \mathbf{W}_{fw})$$

The spatially varying learned branch contributions allow the model to weight closeness and weekly history differently across locations. The physical density estimate for $t+h$ generated at $t$ is then recovered as $\hat{\mathcal{D}}_{future} = \hat{\mathcal{D}}_{t+h\mid t} = s(\hat{Z}_{future} + 1)/2$, which is nonnegative; the scale must cover the intended density range, and the hyperbolic tangent bounds recovered values to $[0, s]$. Per-image normalization is avoided because it would remove count magnitude.

\subsection{Stage 3: Multi-Agent Signal Optimization (RL Control)}
\label{sec:stage3}
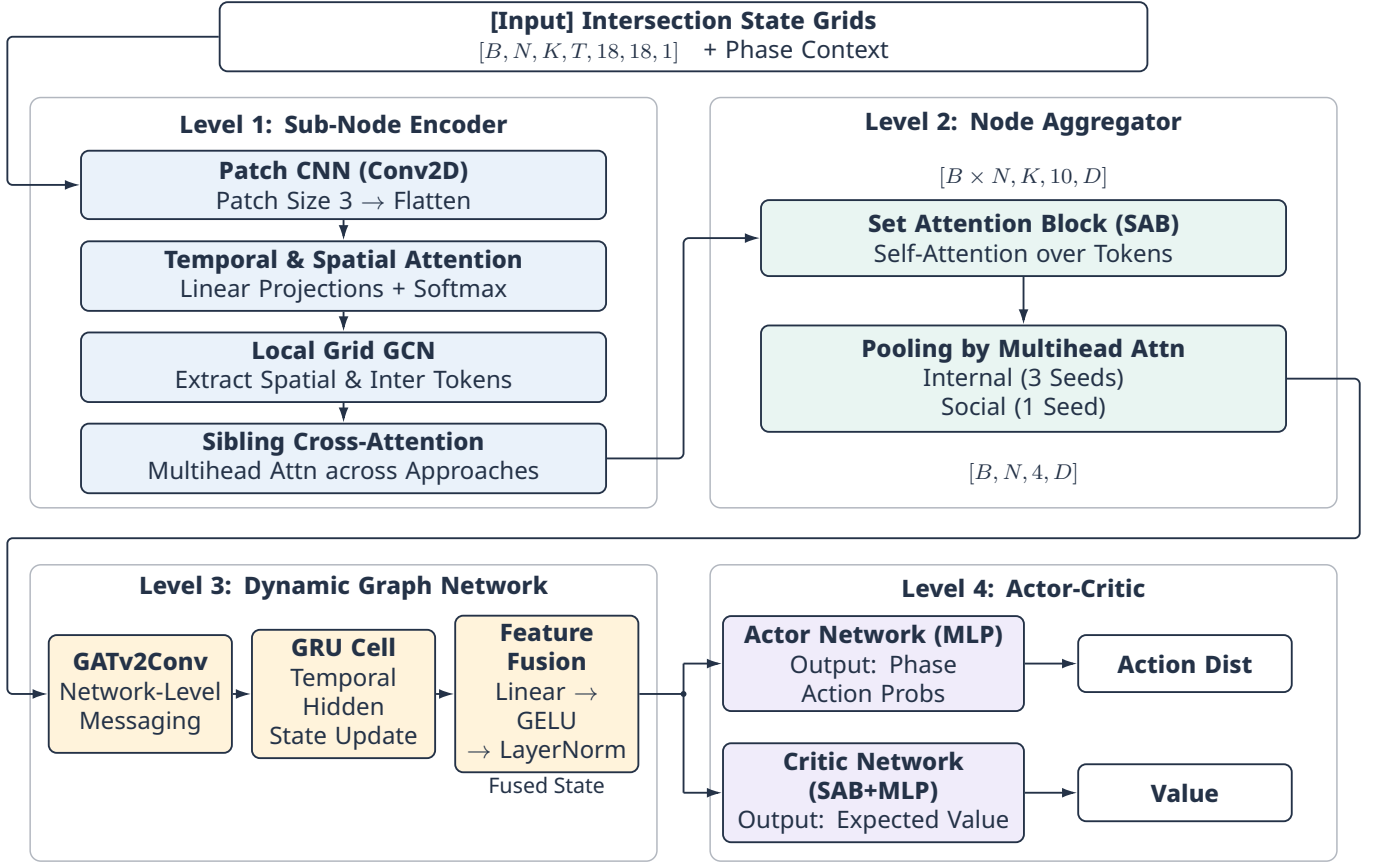
\begin{figure}[!htbp]
\centering
\resizebox{\linewidth}{!}{%
\begingroup\sffamily
\begin{tikzpicture}[font=\sffamily\fontsize{9}{11}\selectfont]
\definecolor{vthreeInk}{HTML}{243247}
\definecolor{vthreeEncoder}{HTML}{EAF2FA}
\definecolor{vthreeAggregator}{HTML}{E9F5F2}
\definecolor{vthreeGraph}{HTML}{FFF3DB}
\definecolor{vthreeHeads}{HTML}{F0ECFA}
\tikzset{
  vthree/frame/.style={draw=vthreeInk!35,line width=0.55pt,
    rounded corners=4pt,fill=white},
  vthree/block/.style={draw=vthreeInk,line width=0.7pt,
    rounded corners=3pt,text=vthreeInk,inner xsep=3pt,inner ysep=3pt,
    align=center,font=\sffamily\fontsize{9}{11}\selectfont},
  vthree/encoder/.style={vthree/block,fill=vthreeEncoder,
    text width=6.7cm,minimum height=0.90cm},
  vthree/aggregator/.style={vthree/block,fill=vthreeAggregator,
    text width=6.7cm,minimum height=1.0cm},
  vthree/graph/.style={vthree/block,fill=vthreeGraph,
    text width=2.2cm,minimum height=1.55cm},
  vthree/head/.style={vthree/block,fill=vthreeHeads,
    text width=3.75cm,minimum height=1.25cm},
  vthree/output/.style={vthree/block,fill=white,
    text width=2.6cm,minimum height=0.75cm},
  vthree/title/.style={text=vthreeInk,text width=7.8cm,align=center,
    font=\sffamily\fontsize{9}{11}\selectfont\bfseries,inner sep=0pt},
  vthree/tensor/.style={text=vthreeInk,align=center,
    font=\sffamily\fontsize{8}{9.5}\selectfont,inner sep=0pt},
  vthree/line/.style={draw=vthreeInk,line width=0.7pt,
    rounded corners=2pt},
  vthree/arrow/.style={vthree/line,-{Latex[length=2mm,width=1.3mm]}}
}

\draw[vthree/frame] (0.05,0.20) rectangle (8.30,-5.20);
\draw[vthree/frame] (9.00,0.20) rectangle (17.25,-5.20);
\draw[vthree/frame] (0.05,-5.95) rectangle (8.30,-9.85);
\draw[vthree/frame] (9.00,-5.95) rectangle (17.25,-9.85);
\node[vthree/title] at (4.175,-0.15) {Level 1: Sub-Node Encoder};
\node[vthree/title] at (13.125,-0.15) {Level 2: Node Aggregator};
\node[vthree/title] at (4.175,-6.25) {Level 3: Dynamic Graph Network};
\node[vthree/title] at (13.125,-6.25) {Level 4: Actor-Critic};

\node[vthree/block,fill=white,text width=12cm,minimum height=0.75cm]
  (vthree-input) at (8.65,1.00)
  {\textbf{[Input] Intersection State Grids}\\
   {\fontsize{8}{9.5}\selectfont $[B,N,K,T,18,18,1]$}\quad + Phase Context};

\node[vthree/encoder] (vthree-patch) at (4.175,-0.95)
  {\textbf{Patch CNN (Conv2D)}\\Patch Size 3 $\rightarrow$ Flatten};
\node[vthree/encoder] (vthree-attn) at (4.175,-2.15)
  {\textbf{Temporal \& Spatial Attention}\\Linear Projections + Softmax};
\node[vthree/encoder] (vthree-gcn) at (4.175,-3.35)
  {\textbf{Local Grid GCN}\\Extract Spatial \& Inter Tokens};
\node[vthree/encoder] (vthree-cross) at (4.175,-4.55)
  {\textbf{Sibling Cross-Attention}\\Multihead Attn across Approaches};

\node[vthree/tensor] at (13.125,-0.85) {$[B\times N,K,10,D]$};
\node[vthree/aggregator] (vthree-sab) at (13.125,-1.65)
  {\textbf{Set Attention Block (SAB)}\\Self-Attention over Tokens};
\node[vthree/aggregator,minimum height=1.4cm] (vthree-pma) at (13.125,-3.50)
  {\textbf{Pooling by Multihead Attn}\\Internal (3 Seeds)\\Social (1 Seed)};
\node[vthree/tensor] at (13.125,-4.75) {$[B,N,4,D]$};

\node[vthree/graph] (vthree-gat) at (1.50,-7.65)
  {\textbf{GATv2Conv}\\Network-Level\\Messaging};
\node[vthree/graph] (vthree-gru) at (4.175,-7.65)
  {\textbf{GRU Cell}\\Temporal Hidden\\State Update};
\node[vthree/graph] (vthree-fuse) at (6.85,-7.65)
  {\textbf{Feature Fusion}\\Linear $\rightarrow$ GELU\\$\rightarrow$ LayerNorm};
\node[vthree/tensor] at (6.85,-8.85) {Fused State};

\node[vthree/head] (vthree-actor) at (11.15,-7.25)
  {\textbf{Actor Network (MLP)}\\Output: Phase Action Probs};
\node[vthree/head] (vthree-critic) at (11.15,-8.95)
  {\textbf{Critic Network (SAB+MLP)}\\Output: Expected Value};
\node[vthree/output] (vthree-action) at (15.25,-7.25) {\textbf{Action Dist}};
\node[vthree/output] (vthree-value) at (15.25,-8.95) {\textbf{Value}};

\draw[vthree/arrow] (vthree-input.west) -- (-0.25,1.00)
  -- (-0.25,-0.95) -- (vthree-patch.west);
\draw[vthree/arrow] (vthree-patch.south) -- (vthree-attn.north);
\draw[vthree/arrow] (vthree-attn.south) -- (vthree-gcn.north);
\draw[vthree/arrow] (vthree-gcn.south) -- (vthree-cross.north);

\draw[vthree/arrow] (vthree-cross.east) -- (8.65,-4.55)
  -- (8.65,-1.65) -- (vthree-sab.west);
\draw[vthree/arrow] (vthree-sab.south) -- (vthree-pma.north);

\draw[vthree/arrow] (vthree-pma.east) -- (17.55,-3.50)
  -- (17.55,-5.60) -- (-0.25,-5.60)
  -- (-0.25,-7.65) -- (vthree-gat.west);
\draw[vthree/arrow] (vthree-gat.east) -- (vthree-gru.west);
\draw[vthree/arrow] (vthree-gru.east) -- (vthree-fuse.west);

\draw[vthree/line] (vthree-fuse.east) -- (8.65,-7.65);
\draw[vthree/line] (8.65,-7.25) -- (8.65,-8.95);
\fill[vthreeInk] (8.65,-7.65) circle[radius=1.1pt];
\draw[vthree/arrow] (8.65,-7.25) -- (vthree-actor.west);
\draw[vthree/arrow] (8.65,-8.95) -- (vthree-critic.west);
\draw[vthree/arrow] (vthree-actor.east) -- (vthree-action.west);
\draw[vthree/arrow] (vthree-critic.east) -- (vthree-value.west);
\end{tikzpicture}
\endgroup%
}
\caption{Hierarchical signal controller with sub-node encoding, set-based aggregation, GATv2 and GRU processing, and phase-policy and state-value heads.}
\label{fig:stage-3}
\end{figure}

The multi-intersection traffic signal control problem (Figure~\ref{fig:stage-3}) is addressed by a proposed hierarchical graph-based MARL controller under the partial-observability formulation in Section~3.2.

\vspace{0.5em}
\noindent\textbf{Sim-to-Real Bridge via Virtual Projection:}

To supply simulated density observations to the controller that are compatible in grid and representation with the Stage~1 output (Section~\ref{sec:stage1}), we propose a projection module (VirtualCamera) using Simulation of Urban MObility (SUMO)~\cite{lopez2018sumo}. Let a vehicle's 3D world coordinate be $P_{world} = [x, y, z, 1]^T$. We map this to a 2D camera pixel coordinate $p_{img} = [u, v, 1]^T$ using standard pinhole camera intrinsics $\mathbf{K}$ and an extrinsic transformation matrix defined by the camera's pitch and heading rotations ($\mathbf{R}_x, \mathbf{R}_z$) and translation $\mathbf{T}$:

$$p_{img} \sim \mathbf{K}[\mathbf{R}_x \mathbf{R}_z | \mathbf{T}] P_{world}$$

Here $\sim$ denotes homogeneous equality up to scale, and $\mathbf{R}_x \mathbf{R}_z$ and $\mathbf{T}$ form the world-to-camera transform under the selected coordinate convention. Projections are retained only when camera depth is positive and image coordinates fall within the image bounds. The image coordinates are then mapped onto the $18 \times 18$ density grid using the image-to-grid scaling described in Section~3.1.2. Let $V_{proj}$ be the number of retained projected vehicles, with scaled grid coordinates $(u_k, v_k)$. A discrete Gaussian with bandwidth $\sigma > 0$ in grid pixels is normalized over the finite density grid $\Omega$:

$$g_k(u,v) = \exp\left(-\frac{(u-u_k)^2+(v-v_k)^2}{2\sigma^2}\right), \qquad
\mathcal{D}_{sim}(u,v) = \sum_{k=1}^{V_{proj}} \frac{g_k(u,v)}{\sum_{(a,b)\in\Omega}g_k(a,b)}$$

Before any perturbations, the spatial sum of $\mathcal{D}_{sim}$ equals $V_{proj}$ within numerical precision; the map is zero when no vehicles are projected. This construction represents in-view projected vehicles without modeling occlusion or density-estimation errors.

\vspace{0.5em}
\textbf{Domain randomization} is proposed for training, following the general idea of varying simulated conditions~\cite{tobin2017domain}. Perturbations to camera calibration and synthetic density maps would be used to explore robustness to projection and density-estimation errors.

\vspace{0.5em}
\noindent\textbf{Hierarchical State Representation:} The local observation for each intersection combines current and historical Stage~1 estimated densities, Stage~2 forecast densities in physical scale after inverse normalization, and signal-phase and mask information supplied by the controller; messages derived from observations at connected intersections are also used. The proposed HierarchicalSignalNet builds representations at the approach, intersection, and network levels, with separate actor and critic heads.

\subsubsection{Sub-Node (Patch) Level}
Each intersection is represented by a single-channel $18\times18$ density map per approach. For the proposed input interface, $T_h$ observed maps through decision $t$ are stacked with one Stage~2 forecast for $t+h$, generated at $t$ from available history and converted back to physical count-density. Thus, $T_{in}=T_h+1$ is the temporal length $T$ in Figure~\ref{fig:stage-3}, distinct from the control horizon in Section~3.2. No future ground truth is used. Here, $B$ is the batch size, $N$ the number of intersections, $K$ the number of padded approach slots, and $D$ the embedding width.

A patch convolution with kernel size and stride 3 yields a $6\times6$ grid of 36 patch features per approach. Linear temporal scores are softmax-weighted and pooled per patch, and linear spatial scores reweight the 36 patches. This pooling has no explicit encoding of within-window order or observed-versus-forecast identity; the later recurrent unit tracks history across decision steps. A local Graph Convolutional Network (GCN) connects horizontal and vertical patch neighbors, using row-normalized neighborhood aggregation, a learned projection, and ReLU; graph convolution provides the general basis for this operation~\cite{kipf2017semi}. Nine $2\times2$ patch regions are pooled into nine spatial tokens, and the whole grid into one interaction token. Sibling multi-head attention updates the interaction tokens across the $K$ approaches of an intersection, masking invalid approaches. The encoder output has shape $[B\times N,K,10,D]$.

\subsubsection{Node (Intersection) Level}
A Set Attention Block (SAB) is applied over valid approach tokens, followed by Pooling by Multihead Attention (PMA) with three learned seeds for spatial tokens and one for interaction tokens, following the Set Transformer formulation~\cite{lee2019set}. Padded approaches are masked in both attention and pooling. The output comprises four tokens per intersection, denoted $U_t^i$, with batched shape $[B,N,4,D]$; its fourth token, $u_t^i$, represents interactions.

\subsubsection{Network Level}
Inter-intersection coordination uses Graph Attention Network v2 (GATv2)~\cite{brody2022attentive} over the directed road graph $\mathcal{E}$, using the interaction tokens as inputs. Neighbor messages are exchanged during execution:
\begin{align*}
m_t^i &= \operatorname{GATv2}(\{u_t^j:j\in\mathcal{V}\},\mathcal{E})_i, \\
h_t^i &= \operatorname{GRU}(m_t^i,h_{t-1}^i), \qquad
z_t^i = \operatorname{Fuse}(U_t^i,h_t^i).
\end{align*}
The Gated Recurrent Unit (GRU)~\cite{cho2014learning} maintains recurrent memory $h_t^i$ across decision steps. Fusion combines the four intersection tokens and recurrent state using a linear projection, Gaussian Error Linear Unit (GELU), and layer normalization, as shown in Figure~\ref{fig:stage-3}.

\subsubsection{Policy Value Heads (Phase-Selection Actor-Critic)}
The actor is a Multilayer Perceptron (MLP) that samples a phase action:
$$a_t^i \sim \pi_\theta(\cdot\mid z_t^i,\phi_t^i,M_t^i), \qquad a_t^i\in\mathcal{A}_t^i.$$
Here, $\phi_t^i$ is the signal-phase context and $M_t^i$ is a vector mask implementing the valid-phase and timing constraints of Section~3.2. Invalid logits are set to negative infinity before softmax, retaining at least one admissible action. During yellow clearance, the previously selected target green is retained.

The centralized critic uses SAB and an MLP to estimate a scalar expected discounted team return from the collection of intersection features:
$$V_\varphi(Z_t), \qquad Z_t=\{z_t^i:i\in\mathcal{V}\}.$$
The whole-network critic is used only during training; each actor uses its local features, phase and timing information, and neighbor messages. Advantages use Generalized Advantage Estimation (GAE)~\cite{schulman2016high}, with $\lambda=0.95$ as a proposed setting.

\subsubsection{Multi-Objective Reward Function}
The base reward is augmented with fairness and coordination terms:
$$R_t=-\alpha\sum_{i=1}^{N}(W_{t+1}^i-W_t^i)-\beta\sum_{i=1}^{N}\operatorname{QueueVar}_t^i-\eta\sum_{(i,j)\in\mathcal{E}}P_t^{ij}.$$
Here, $W_t^i$ is defined in Section~3.2, and $\operatorname{QueueVar}_t^i$ is the variance of SUMO halting-vehicle counts across the valid approaches of intersection $i$, used as a training-time reward signal. The nonnegative coefficients $\alpha$, $\beta$, and $\eta$ are fixed weights to be selected; $\gamma$ remains the discount factor. The coordination term is a proposed downstream congestion proxy:
$$P_t^{ij}=\max(0,\hat{N}_{t+h\mid t}^{ij}-C_{ij}), \qquad
\hat{N}_{t+h\mid t}^{ij}=\sum_{(u,v)\in\Omega_{ij}}\hat{\mathcal{D}}_{t+h\mid t}(u,v).$$
The inverse-scaled Stage~2 forecast $\hat{\mathcal{D}}_{t+h\mid t}$ is summed over the receiving-road region $\Omega_{ij}$ on a single camera grid for link $i\rightarrow j$. The threshold $C_{ij}$ is a congestion threshold in vehicle counts for that same region. Region masks and thresholds require calibration. Density represents vehicle presence, so this proxy does not directly measure queued vehicles, spillback, or delay.

\subsubsection{Training Algorithm and Operational Constraints}
The proposed training procedure uses Proximal Policy Optimization (PPO) with a clipped surrogate objective and entropy bonus in a Multi-Agent Proximal Policy Optimization (MAPPO) framework~\cite{yu2022surprising}, with actor neighbor communication and a centralized critic. The phase-only minimum-green mask follows Section~3.2: elapsed green time excludes yellow, the environment inserts yellow clearance, and additional phase changes are blocked during clearance.

\section{Results \& Discussion}

\subsection{Stage 1: Vehicle Density Estimation}

Given the limited dataset size (40 samples), we employ a rigorous 5-fold cross-validation strategy with ensemble averaging to maximize generalization. Each fold trains with 10$\times$ oversampling via stochastic augmentation (horizontal flip, brightness/contrast jitter, saturation perturbation, Gaussian noise injection, random blur, and random crop-resize). Training uses AdamW ($\text{lr} = 2\times10^{-4}$, weight decay $10^{-3}$) with cosine annealing over 600 epochs and early stopping (patience=120) based on validation MAE. The composite loss combines pixel-level MSE with a count regularization term ($\lambda_c = 2.0$):

$$\mathcal{L} = \frac{1}{HW}\sum_{u,v}(\hat{\mathcal{D}}_{u,v} - \mathcal{D}^*_{u,v})^2 + \lambda_c \left(\sum_{u,v}\hat{\mathcal{D}}_{u,v} - N^*\right)^2$$

\noindent where $N^*$ is the ground-truth vehicle count. Gradient clipping at 5.0 stabilizes training on the small dataset.

\begin{table}[h]
\centering
\caption{5-fold cross-validation results across three training rounds on 40 real-world intersection images (vehicle counts 12--49). \textbf{Mean Fold MAE} is the mean validation MAE across five folds. The \textbf{Ensemble} columns are computed over the full 40-sample dataset, where four of the five ensemble members have seen each image during training, and are therefore in-sample rather than held-out.}
\begin{tabular}{@{}lcccc@{}}
\toprule
\textbf{Configuration} & \textbf{Mean Fold MAE} & \textbf{Ensemble MAE} & \textbf{Ensemble MAPE} & \textbf{Max Error} \\ \midrule
Round 1 (single model, fixed $\sigma$)  & 2.84 & --- & --- & --- \\
Round 2 (5-fold, fixed $\sigma=1.5$)    & 3.07 $\pm$ 1.03 & 1.83 & 9.2\% & 8.7 \\
Round 3 (5-fold, adaptive $\sigma$)     & 3.32 $\pm$ 0.95 & \textbf{1.44} & \textbf{6.7\%} & \textbf{5.1} \\ \bottomrule
\end{tabular}
\label{tab:stage1}
\end{table}

\begin{table}[h]
\centering
\caption{Per-fold breakdown for Round 3.}
\begin{tabular}{@{}lccc@{}}
\toprule
\textbf{Fold} & \textbf{Best MAE} & \textbf{Best MAPE} & \textbf{Early Stop Epoch} \\ \midrule
Fold 0 & 3.62 & 10.8\% & 252 \\
Fold 1 & 2.00 & 7.7\%  & 158 \\
Fold 2 & 2.67 & 13.8\% & 404 \\
Fold 3 & 3.48 & 18.5\% & 251 \\
Fold 4 & 4.82 & 21.5\% & 203 \\ \midrule
\textbf{Mean} & \textbf{3.32 $\pm$ 0.95} & \textbf{14.5\%} & --- \\ \bottomrule
\end{tabular}
\label{tab:stage1-folds}
\end{table}

The mean fold validation MAE is \textbf{3.32 $\pm$ 0.95 vehicles} (MAPE 14.5\%). Reported scores use validation folds that also guide checkpoint selection.

\vspace{0.5em}
We additionally report an ensemble score of 1.44 vehicles (MAPE 6.7\%, maximum error 5.1), obtained by averaging the predicted density maps of all five fold models over the complete 40-sample dataset. This value must be interpreted with care. Under 5-fold cross-validation each image belongs to the training split of four of the five ensemble members, so this evaluation is predominantly in-sample. The apparent 57\% reduction relative to the mean fold error therefore reflects memorization considerably more than genuine variance reduction, and 1.44 should not be read as a generalization result. A leakage-free ensemble estimate is not obtainable under this protocol, since only a single member is held out per image. The round-to-round comparisons drawn from the ensemble columns inherit the same bias and are reported for completeness rather than as evidence of improvement.

\vspace{0.5em}
Residual errors concentrate on extremely low-count images (12 vehicles, errors of 4--5) due to mean-regression bias inherent to small datasets. The $18\times18$ density map preserves the spatially resolved queue structure per approach, enabling downstream modules to distinguish congested from free-flowing lanes.

\section{Future Work}
Several critical avenues remain open for extending the proposed framework towards city-scale deployment:

\vspace{0.5em}
\noindent\textbf{Dataset Expansion and End-to-End Pipeline Training:} The current evaluation is bottlenecked by a limited dataset, which restricts the full optimization of the vehicle density estimator (Stage 1). A primary immediate objective is to significantly scale the acquisition and annotation of high-resolution traffic frames across diverse intersection topologies, weather conditions, and times of day. Recorded traffic sequences are needed for Stage 2 forecasting, while simulation rollouts are needed for Stage 3 policy training and evaluation.

\vspace{0.5em}
\noindent\textbf{Domain Adaptation and Sim-to-Real Evaluation:} The proposed VirtualCamera projection module and domain randomization require evaluation with calibrated real-world cameras. Future work will explore unsupervised domain adaptation to fine-tune the Stage 1 density regressor directly on target camera distributions. Future work will also evaluate transfer of the phase-only controller after training in SUMO.

\vspace{0.5em}
\noindent\textbf{City-Scale Signal Coordination:} Future work will assess the computational scaling of the proposed graph-based controller (GATv2 with a GRU) on city-level networks comprising hundreds of nodes. One possible approach is to introduce an additional topological hierarchy: clustering junctions into localized traffic districts, and training a meta-agent to manage inter-district coordination while local agents handle intra-district signal coordination.

\vspace{0.5em}
\noindent\textbf{Multi-Modal Traffic Integration:} The current Stage 1 model is optimized primarily for motorized vehicle density. To address the extreme heterogeneity of traffic in developing nations like India, the CNN could be expanded to output class-specific density estimates for pedestrians, two-wheelers, and heavy vehicles. Within the POMDP formulation, future work will investigate reward and constraint designs for pedestrian service and evaluate mixed-traffic safety.

\section{Summary \& Conclusion}
This paper presents the design and partial validation of a three-stage vision-to-control pipeline for adaptive traffic signal control. The primary contributions are:

\begin{itemize}
    \item A compact CNN density estimator evaluated on a dataset of 40 real-world Indian intersection images, achieving a mean five-fold validation MAE of \textbf{3.32 $\pm$ 0.95 vehicles} (MAPE 14.5\%) using adaptive Gaussian kernel density maps. An in-sample five-fold ensemble score of 1.44 vehicles is additionally reported in Section 4.1, but it is not a generalization estimate and is not claimed as one.
    \item A proposed spatio-temporal density-map forecasting module (ST3DNet) and a proposed hierarchical multi-agent RL controller (MAPPO with GATv2 and GRU), with training and evaluation left as future work.
    \item A proposed VirtualCamera projection module with domain randomization, intended to connect simulated density observations to the controller; real-world transfer requires evaluation.
\end{itemize}

\vspace{0.5em}
The Stage 1 results support the feasibility of vehicle counting from $72\times72$ model input images on the evaluated dataset. Training and evaluating Stage 2 on real traffic sequences and Stage 3 on SUMO rollouts remain the next steps toward a deployable system.

\end{document}